\documentclass[sigconf]{acmart}

\newcommand{\bulitem}[1]{\noindent\textbf{#1}}
\renewcommand\footnotetextcopyrightpermission[1]{}
\usepackage{multirow}
\usepackage{amsmath}

\usepackage{amssymb}
\usepackage{enumitem}
\usepackage{algorithm}
\usepackage[noend]{algpseudocode}

\AtBeginDocument{%
  }

\begin{document}

\title{Read, Diagnose and Chat: Towards  Explainable and Interactive LLMs-Augmented Depression Detection in Social Media}




\author{Wei Qin}
\affiliation{%
  \institution{Hefei University of Technology}
  \streetaddress{1 Th{\o}rv{\"a}ld Circle}
  \city{Hefei}
  \country{China}}
\email{qinwei.hfut@gmail.com}

\author{Zetong Chen}
\affiliation{%
  \institution{University of Science and Technology of China}
  \streetaddress{1 Th{\o}rv{\"a}ld Circle}
  \city{Hefei}
  \country{China}}
\email{dvdx@mai.ustc.edu.cn}

\author{Lei Wang}
\affiliation{%
  \institution{Singapore Managerment University}
  \streetaddress{1 Th{\o}rv{\"a}ld Circle}
  \city{Singapre}
  \country{Singapore}}
\email{lei.wang.2019@phdcs.smu.edu.sg}

\author{Yunshi Lan}
\affiliation{%
  \institution{East China Normal University}
  \streetaddress{1 Th{\o}rv{\"a}ld Circle}
  \city{Shanghai}
  \country{China}}
\email{lei.wang.2019@phdcs.smu.edu.sg}

\author{Weijieying Ren}
\affiliation{%
  \institution{Pennsylvania State University}
  \streetaddress{1 Th{\o}rv{\"a}ld Circle}
  \city{Pennsylvania}
  \country{USA}}
\email{-}

\author{Richang Hong}
\affiliation{%
  \institution{Hefei University of Technology}
  \streetaddress{1 Th{\o}rv{\"a}ld Circle}
  \city{Hefei}
  \country{China}}
\email{hongrc.hfut@gmail.com}
\begin{abstract}
  %
Depression detection based on social media content has received increasing attention, as it allows for early diagnosis before the user's psychological state deteriorates. 
Although traditional methods of depression detection can provide a classification of whether the user is depressed or not, they cannot provide human-like explanations and interactions. 
In this paper, we propose a next-generation paradigm for depression detection, namely an interpretable and interactive depression detection system based on LLMs. 
The proposed system not only yields a final diagnosis result, but also offers diagnostic evidence grounded in established diagnostic criteria. Furthermore, it enables users to engage in natural language dialogue with the system, facilitating a more personalized understanding of their mental state based on their social media content. The interactive dialogue allows for the provision of tailored recommendations, which users can utilize to enhance their well-being. In constructing the entire system, we also addressed some non-trivial challenges. 
First, we introduced the chain of thoughts technique and professional depression diagnostic criteria when constructing the prompts, enabling our system to make decisions based on professional diagnosis criteria and provide explanations. Secondly, LLMs are incapable of processing excessively long contextual texts, and the accumulated posts of a single user may amount to tens of thousands of words. To overcome this limitation, we integrated a tweet selector that selects the part of posts for diagnosis. The experiments demonstrate that our depression detection system achieves the best performance across various settings, including full data setting, few-shot setting, zero-shot setting, independent-identical-distribution(IID) setting, and out-of-distribution(OOD) setting. Additionally, case studies reveal the explanation and interactivity of our system.
\end{abstract}




\maketitle

\section{Introduction}


Mental health disorders are a significant and widespread public health concern, with depression affecting an estimated 300 million people worldwide~\cite{WHO2012_depression}. Depression is a leading cause of suicide and ranks as the third leading cause of death for individuals between the ages of 10 and 24 years old. Despite the severity of the issue, depression remains underdiagnosed and undertreated~\cite{allan2014depression, sheehan2004depression}. More than half of those who suffer from depression do not receive any treatment~\cite{WHO2012_depression}. 
\citet{williams2017undiagnosed}
used a depression diagnostic test instrument to discover that 8\% of the population had symptoms and a diagnosis of depression, while 7.6\% had symptoms but had not been diagnosed. Individuals affected by mental disorders often hesitate to seek professional help~\cite{edwards2007reluctance}. However, as more people turn to social media to discuss their struggles and seek emotional support, automatic processing of social media data could be a crucial tool in identifying changes in mental health status before they lead to more serious health consequences.

\begin{figure*}[t]
\begin{center}
\vspace{-4mm}
\includegraphics[width=1.0\textwidth]{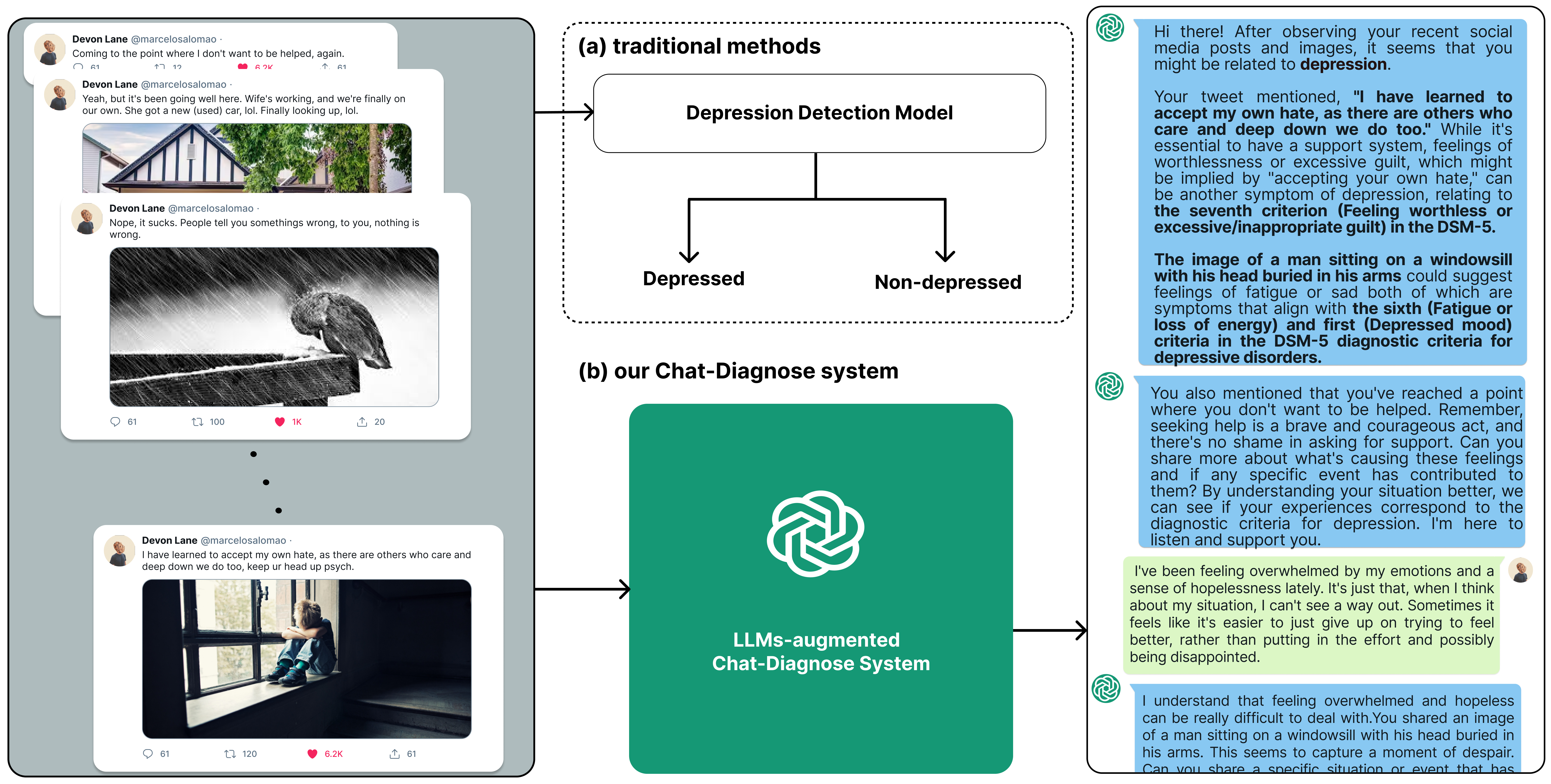}
\end{center}
\vspace{-5mm}
\caption{User posts list is the input of the task. The sub-figure (a) shows that traditional methods only output the final diagnosis result. Our method in sub-figure (b) could provide a diagnosis result, diagnosis evidence and interactive dialogue. }
\vspace{-3mm}
\label{fig:illutration}
\end{figure*}

Currently, many researchers are focusing on the task of detecting depression in social media. To this end, various studies, including those conducted by ~\cite{gui2019twitter, Bucur2023timedep}, have investigated the use of different network structures such as Convolutional Neural Networks (CNNs), Long Short-Term Memory (LSTM) networks, and transformers. Additionally, ~\cite{an2020multimodal} introduced a topic-enriched auxiliary task that improves the detection ability of depression symptoms by enabling the model to better comprehend social media content. Furthermore, ~\cite{cheng2022multimodal} considered the timing of social media posts. 
Notwithstanding their impressive performance, as depicted in Figure ~\ref{fig:illutration}(a), existing methods for depression classification suffer from a crucial limitation: they solely produce binary classification outputs (i.e., depressed or non-depressed) without providing any accompanying explanation. 
This drawback renders the classification results less persuasive to users, and provides limited reference information for medical practitioners as well.
Another primary constraint is their pool interactivity and lack of feedback mechanisms. 
%
An interactive environment is necessary to allow users to express their mental state more deeply and to obtain effective personalized recommendations.
However, little research has been conducted in this domain.

Large language models (LLMs), such as GPT-3 ~\cite{GPT_3_NEURIPS2020_1457c0d6}, OPT ~\cite{Zhang2022OPTOP}, and PaLM ~\cite{Chowdhery2022PaLMSL}, have evolved rapidly and have demonstrated remarkable performance across various natural language processing (NLP) tasks. As LLMs continue to expand in size and training data, they are strong semantic understanding and reasoning abilities~\cite{wei2022emergent}. The emergence of LLMs has brought about a paradigm shift in research. Previously, applying models to downstream tasks typically involved adjusting model parameters through backpropagation. However, the latest development of LLMs ~\cite{touvron2023llama} has enabled both researchers and practitioners to facilitate learning during the forward process by constructing prompts, namely In-Context Learning (ICL)
%
%
Recently, LLMs with ICL paradigm are leveraged to solve many tasks including movie recommendation ~\cite{Gao2023ChatRECTI}, medical image classification ~\cite{wang2023chatcad} and document information extraction ~\cite{he2023icld3ie}. However, to date, there has been no investigation into the ability of LLMs to handle the depression detection task. 

In this work, we propose a next-generation paradigm for depression detection in social media, namely an explainable and interactive LLMs-augmented depression detection system---\textit{Chat-Diagnose}. 
As illustrated in Figure ~\ref{fig:illutration}(b), Chat-Diagnose system combines established diagnostic criteria with social media content to assess the risk of depression in users and generate interpretable diagnostic evidence. Following the diagnostic process, Chat-Diagnose system communicates the diagnosis results to users through natural dialogue, along with relevant diagnostic evidence. 
In this interactive dialogue, users can describe their mental state more deeply through our guided questioning based on their social media content, and we can more accurately assess their mental state and provide appropriate personalized recommendations based on the content of the conversation.

However, several main challenges arise when applying LLMs to depression detection in social media. 
The first challenge is how to force the system to provide explanations based on professional diagnostic knowledge. To address this challenge, 
we first incorporate external professional diagnostic knowledge (DSM depression diagnosis criteria) as part of the prompt.
On top of this, we introduce the chain of thoughts(CoT) technique and construct demonstrations, which enable our LLMs-augmented system to make diagnoses based on professional diagnosis criteria and provide diagnostic evidence. Figure ~\ref{fig:dsm} shows the DSM depression diagnosis criteria and the demonstrations.
The second challenge lies in the fact that users may have hundreds or thousands of posts, and the lengthy context can overwhelm LLMs. Therefore, we build a tweet selector module to screen each post and input the most related ones to LLMs. 
%

\begin{figure*}[t]
\begin{center}
\vspace{-4mm}
\includegraphics[width=1.0\textwidth]{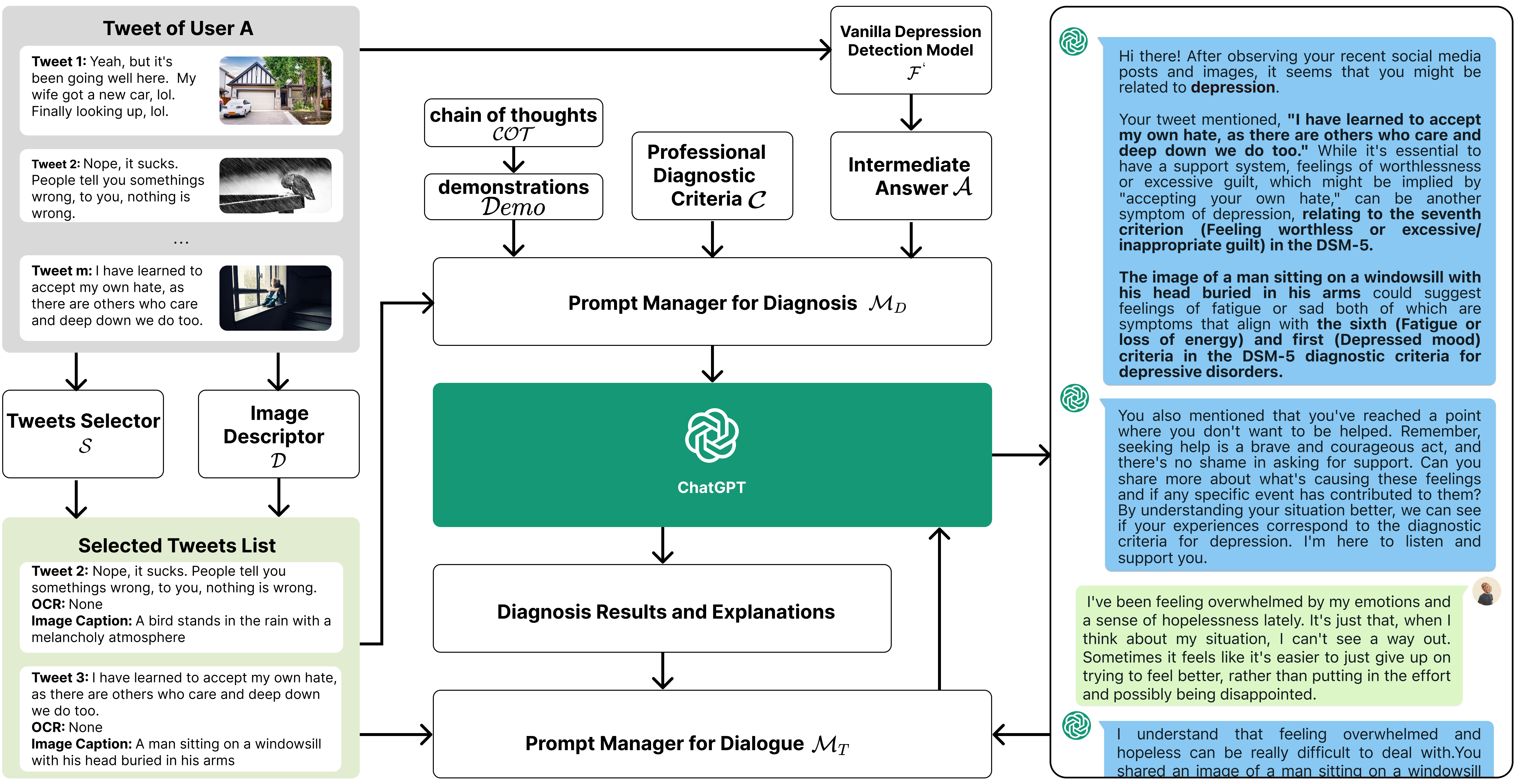}
\end{center}
\vspace{-4mm}
\caption{The framework of our Chat-Diagnose system.}
\label{fig:framework}
\end{figure*}

Our Chat-Diagnose system can operate in zero-shot, few-shot and full-training settings. In the zero-shot setting, no examples are provided to our system. In the few-shot setting, our system inputs only two CoT demonstrations. In the full-training setting, our method additionally inputs the prediction outputs of a full-training vanilla depression detection model as answer heuristics to LLMs.
As the currently most available LLMs cannot process images, we utilize image captioning and OCR to convert image content into textual descriptions. 
%

We conducted extensive experiments on Twitter dataset and Weibo dataset. Our approach achieved state-of-the-art performance in multiple zero-shot, few-shot, and full-data settings, as well as independent-identical-distribution (IID) and out-of-distribution (OOD) test data. In addition, ablation studies confirmed the effectiveness of each module in our system. Case studies demonstrated the quality of the explanation and interactivity provided by our system.

Our contributions are summarized as follows: 1). We developed a new depression detection system(Chat-Diagnose) around LLM that provides diagnostic evidence and personalized advice through natural language dialogue. 2). By introducing CoT technology and a tweets selector, we improve explainability and avoid LLMs' limitations in handling excessive posts. 3). Extensive experiments and case studies demonstrate our system's superior performance, explainability, and interactivity in various settings.

%

\section{Related Work}

\bulitem{Depression detection in social media.} 
Early research on depression detection focused on analyzing language cues identified in psychology literature, such as self-focused language indicated by the frequent use of the pronoun "I" ~\cite{rissola2020beyond_37,rude2004language_38} and dichotomous thinking expressed through absolute words like "always" and "never" ~\cite{fekete2002internet_18}. Reece et al. ~\cite{reece2017instagram_35} demonstrated that individuals with depression tend to post images with sadder, less happy, and darker tones compared to healthy individuals. With the advent of deep learning, various neural network architectures, such as CNNs ~\cite{yates2017depression_58}, LSTMs ~\cite{skaik2020using_43,trotzek2018utilizing_49}, and transformers ~\cite{Bucur2023timedep,sawhney2020time_40}, have been explored to detect depression in social media and have achieved remarkable performance. Gui et al. ~\cite{gui2019twitter} proposed a cooperative multi-agent reinforcement learning approach using two agents to select relevant textual and visual information for classification. An et al. ~\cite{an2020multimodal} proposed a multimodal topic-enriched auxiliary learning approach that improves the primary task of depression detection through auxiliary tasks on visual and textual topic modeling. Bucur et al. ~\cite{Bucur2023timedep} used time-adapted weights ~\cite{chui_time_10} or Time-Aware LSTMs ~\cite{cheng2022multimodal,sawhney2020time_40} to incorporate the time component of data for mental health problem detection.

Deep learning methods achieve dominating performance but are limited by explanation and interactivity. High-quality explanations can convince users to seek professional medical help and increase the likelihood of recovery, and allow doctors to screen users more effectively. The system's interactivity enables users to provide detailed descriptions of their condition and receive personalized advice. We refer to depression detection systems with high explanation and interactivity as a next-generation paradigm for depression detection in social media.

\bulitem{In-Context Learning.} 
Large Language Models (LLMs), including GPT-3~\cite{GPT_3_NEURIPS2020_1457c0d6}, OPT~\cite{Zhang2022OPTOP}, and PaLM~\cite{Chowdhery2022PaLMSL}, are known for their impressive emergent abilities that scale with model and corpus size~\cite{wei2022emergent_37}. These abilities are learned from demonstrations containing only a few examples in the context, which is known as in-context learning~\cite{dong2022survey_8}. To enable reasoning in LLMs, a few-shot prompting strategy called Chain-of-Thought (CoT) prompting has been proposed by 
\citet{wei2022chain_38},
which involves adding multiple reasoning steps to the input question. Recent works\cite{wang2022towards_33, Suzgun2022ChallengingBT_31, Shaikh2022OnST_30} have aimed to improve CoT prompting in various aspects, such as prompt format~\cite{chen2022program_4}, prompt selection~\cite{lu2022dynamic_24}, prompt ensemble~\cite{wang2022self_36}, and problem decomposition~\cite{zhou2023leasttomost_46}. Despite being initially developed for NLP tasks, LLMs with in-context learning have been found to exhibit few-shot or zero-shot abilities for multimodal problems, such as visual question-answering tasks~\cite{yang2022empirical_42, cao-etal-2022-prompting, zeng2022socratic_43}. Additionally, pre-trained models have shown promising few-shot performance in vision-and-language tasks, as demonstrated by Frozen~\cite{tsimpoukelli2021multimodal_32}. However, to the best of our knowledge, this is the first work to explore the use of LLMs with in-context learning for depression detection.

\section{Method}

\subsection{Problem Setup}
Depression detection in social media refers to the diagnosis of whether a user is depressed or not based on their social media content, which may include both text and images. Specifically, we represent each user sample as $(x, y)$, where $x$ is a list of posts from the user's social media account, including multiple posts ${x_{t_1}, x_{t_2}, ..., x_{t_m}}$. Each post $x_{t}$ contains both text $x_{t}^{text}$ and picture $x_{t}^{pic}$. $y$ indicates whether the user has been diagnosed as a depressed individual or not. The traditional depression detection system, denoted by $\mathcal{F}^{*}$, utilizes the user's social media content to determine whether the user exhibits symptoms of depression:
\begin{equation}
    \hat{y} = \mathcal{F}^{*}(x) = \mathcal{F}^{*}({x_{t_1}, x_{t_2}, ..., x_{t_m}})
\end{equation}
where $\hat{y}$ denotes the predicted diagnosis result whether the user is depressed or not.

In this paper, we propose a novel but more challenging paradigm for the depression detection system. The system $\mathcal{F}$ is required not only to provide detection results but also to provide interpretability (diagnostic evidence) and an interactive dialogue system.
\begin{equation}
    \hat{y}, \mathcal{E} = \mathcal{F}(x) = \mathcal{F}({x_{t_1}, x_{t_2}, ..., x_{t_m}})
\end{equation}
where $\hat{y}$ and $\mathcal{E}$ denotes the diagnosis results and the generated diagnosis explanation. Moreover, the system is equipped with an interactive chat module that proactively engages with users. This module can be represented as:
\begin{equation}
    R_t =  \mathcal{F}(c_t, h_{t-1},\hat{y}, \mathcal{E}, x)
\end{equation}
where $c_t$ denotes the input provided by the user at round $t$ and $R_t$ denotes the response generated by the system at round $t$.  $h_{t-1} = {R_1, c_2, R_2, ..., R_{t-1}, c_{t-1}}$ represents the dialogue history up to and including the previous t-1 rounds. Here, $h_0 = None$.

\subsection{Depression Detection System}
\bulitem{Tweet Selector $\mathcal{S}$: } 
The tweet selector is introduced to filter out excessive posts from the user, since LLMs cannot process too long text. The tweet selector $\mathcal{S}$ will select $n$ posts as the input LLMs:
\begin{equation}
    {x_{t^{'}_1}, x_{t^{'}_2}, ... , x_{t^{'}_n}} = \mathcal{S}({x_{t^{}_1}, x_{t^{}_2}, ... , x_{t^{}_m}})
\end{equation}
where ${x_{t^{}_1}, x{t^{}_2}, ... , x{t^{}_m}}$ represents the collection of all $m$ posts from a particular user, while ${x_{t^{'}_1}, x_{t^{'}_2}, ... , x_{t^{'}_n}}$ denotes the subset of n posts that are selected.
We experiment with three tweet selectors: the random selector, the recent selector and the sentiment selector. The random selector randomly selects $n$ posts, while the recent selector selects the $n$ most recent posts. The sentiment selector utilizes a sentiment analysis model to rank all $m$ posts based on their negativity scores, and then select the $n$ most negative posts. We employ GPT-3.5 as our sentiment analysis model.


\bulitem{Image Descriptor $\mathcal{D}$:}
The present module employs Image Caption and Optical Character Recognition (OCR) technology to transform images into textual information, which is a crucial feature as many Language and Learning Models (LLMs) are only capable of processing text. One common approach to extract semantics from images and represent it using textual descriptions is through image captioning ~\cite{cao-etal-2022-prompting,yang2022}. The process starts with extracting the text from the image using open-source Python packages, such as EasyOCR2. Subsequently, a pre-trained image captioning model named ClipCap ~\cite{mokady2021clipcap} is applied. This model is particularly useful for generating high-quality captions for low-resolution web images. The captions produced by ClipCap typically describe the most prominent objects or events depicted in the image. The left part of Figure ~\ref{fig:framework} demonstrates the effect of the Image Descriptor, which utilizes image caption and OCR to transform the visual information of an image into textual form.
In the future, with LLMs capable of handling images such as GPT-4, we can directly input both images and text into the model without the need for this intermediary step.

\bulitem{Professional Diagnostic Criteria $\mathcal{C}$:}
\begin{figure}[]
\begin{center}
\includegraphics[width=0.46\textwidth]{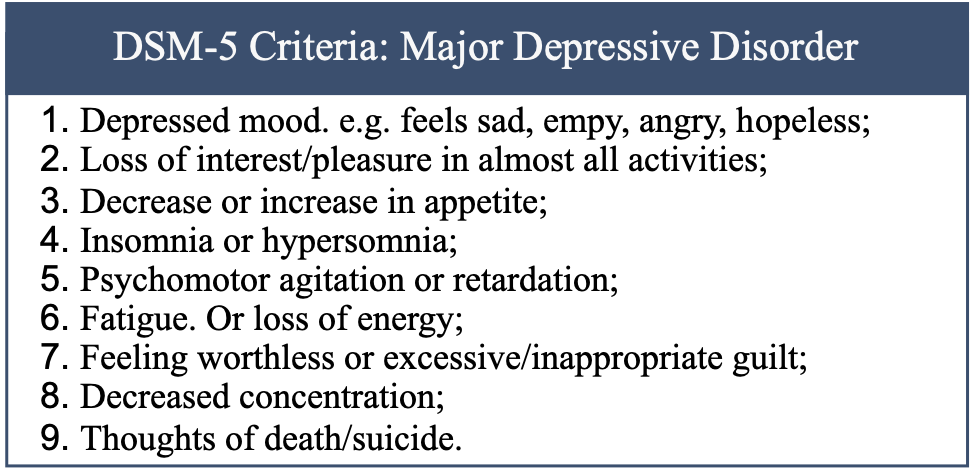}
\end{center}
\vspace{-4mm}
\caption{The abbreviated version of the DSM diagnostic criteria for major depressive disorder is provided, with reference to the complete version available in ~\cite{Ma2021_DSM}.}
\vspace{-5mm}
\label{fig:dsm}
\end{figure}
To facilitate the detection system's ability to perform diagnostic procedures based on professional knowledge, rather than just classification, we have integrated the DSM depression diagnostic criteria as the Professional Diagnostic Criteria $\mathcal{C}$, as depicted in Figure ~\ref{fig:dsm}. The Diagnostic and Statistical Manual of Mental Disorders (DSM) serves as the primary reference for psychiatric diagnoses in the United States. 

\bulitem{Demonstrations $\mathcal{D}emo$:} 
In-context learning usually involves introducing a few ($k$) demonstrations to enable the model to learn how to accomplish the given task. 
Specifically, a demonstration refers to reformulating a labeled sample into a question-answer pair format in a text-based representation that can be understood by LLMs.
Besides, utilizing few demonstrations can effectively constrain LLMs, enabling them to output results in a fixed format~\cite{dong2022survey_8}.

\bulitem{Chain-of-Thought $\mathcal{COT}$:}
The Chain of Thoughts (CoT) technique involves incorporating reasoning steps into demonstrations ($\mathcal{D}emo$) to obtain the final results. This technique enables LLMs to detect depression and provide diagnostic evidence based on the Professional Diagnostic Criteria $\mathcal{C}$. The CoT demonstration comprises triples, which include input, output, and reasoning steps. The input includes the question, options, and posts, while the output indicates the final answer (diagnosis). The reasoning steps in the chain of thought reveal how to derive the answer from the question. The CoT demonstration follows a specific organizational format:


\begin{itemize}[label={}, labelsep=0pt, leftmargin=10pt]
    \item \textbf{Question:} \texttt{Analyze someone's tweets using DSM-5 to determine if they are depressed.}
    \item \textbf{Options:} \texttt{[A.depressed B.non-depressed]}.
    \item \textbf{Posts:} \texttt{[Posts of User]}.
    \item \textbf{Answer:} \texttt{[A or B]}.
    \item \textbf{Explanation:} \texttt{[Reasoning steps]}.
\end{itemize}

\bulitem{Vanilla Depression Detection Model $\mathcal{F}^{*}$ and Answer Heuristics $\mathcal{A}$:} 
In the full data setting, to fully exploit all available training samples, a conventional depression detection model $\mathcal{F}^{*}$ is introduced and trained on the entire training dataset. The resulting prediction probabilities from this model are utilized as answer heuristics $\mathcal{A}$ and subsequently fed into the LLM as inputs. This can be expressed mathematically as:
\begin{align}
{p}(A), {p}(B) &= \mathcal{F}^{*}({x_{t_1}, x_{t_2}, ..., x_{t_m}}) \\
\mathcal{A} &= \{{p}(A), {p}(B)\}
\end{align}
where ${p}(A)$ and ${p}(B)$ are the predicted probabilities that the test user is depressed or not depressed by the conventional depression detection model, respectively, and are considered as the answer heuristics $\mathcal{A}$. 
Specifically, the Time2VecTransformer ~\cite{Bucur2023timedep} is adopted as the vanilla depression detection model. However, in the few-shot and zero-shot settings, the LLM is not provided with any answer heuristics.

\begin{algorithm}
\caption{Algorithm of our depression detection system}
\begin{algorithmic}[1]
\Statex \textbf{Step 1: detect the depression}
\Statex \textbf{Input:} $x$, $\mathcal{COT}$-$\mathcal{D}emo$, $\mathcal{C}$, $\mathcal{L}$ \Comment{$\mathcal{L}$ denotes LLMs}
\Statex \textbf{Output:} $\hat{y}$, $\mathcal{E}$
\Procedure{$\mathcal{F}$}{ $x$, $\mathcal{COT}$-$\mathcal{D}emo$, $\mathcal{C}$} 
\State $\mathcal{A} \gets \mathcal{F}^{*}(x)$
\State $x^{D} \gets \mathcal{D}(x)$
\State $x^{DS} \gets \mathcal{S}(x^{D})$
\State prompt$_D \gets \mathcal{M}_{D}(x^{DS},\mathcal{A}, \mathcal{COT}$-$\mathcal{D}emo)$ 
\State $\hat{y}, \mathcal{E} \gets \mathcal{L}($ prompt$_D)$ 
\EndProcedure 
\Statex 
\Statex \textbf{Step 2: interactive dialogue}
\Statex \textbf{Input:} $x$, $\hat{y}$, $\mathcal{E}$, $\mathcal{L}$
\Procedure{$\mathcal{F}$}{$x$, $\hat{y}$, $\mathcal{E}$}
\State DialogueShow($\hat{y}$, $\mathcal{E}$)
\State $c \gets None$
\State $h \gets None$
\While {dialogue ends}
\State prompt$_T \gets \mathcal{M}_{T}(c, x^{DS}, \hat{y}, \mathcal{E}, h)$ 
\State $R \gets \mathcal{L}$(prompt$_T$)
\State $h \gets h + R$
\State DialogueShow($R$)
\State $c \gets$ user input in this round
\State $h \gets h + c$
\EndWhile
\EndProcedure
\end{algorithmic}
\label{alg:search}
\end{algorithm}

\bulitem{Prompt Manager for Diagnosis $\mathcal{M}_{D}$:} The prompt manager has been designed to convert all clues into a language that LLM can understand. Specifically, the prompt manager for diagnosis $\mathcal{M}_{D}$ is responsible for constructing prompts that instruct LLMs on how to diagnose a user. We conduct the prompt with the question, the $\mathcal{CoT}$ demonstrations, the professional diagnostic criteria $\mathcal{C}$ and the answer heuristics $\mathcal{A}$:

\begin{itemize}[label={}, labelsep=0pt, leftmargin=10pt]
    \item \textbf{Question:} \texttt{Analyze someone's tweets using DSM-5 to determine if they are depressed.}
    \item \textbf{Options:} \texttt{[A.depressed B.non-depressed]}.
    \item \textbf{Diagnosis criteria:} \texttt{[DSM diagnosis criteria]}
    \item \textbf{Demonstrations:} \texttt{[$\mathcal{COT}$ Demonstrations]}.
    \item \textbf{Posts:} \texttt{[Posts of User]}.
    \item \textbf{Answer candidate:} \texttt{[A P(A)]; [B P(B)]}.
    \item \textbf{Answer:} 
    \item \textbf{Explanation:} 
\end{itemize}
where \texttt{$\mathcal{COT}$ Demonstrations} refers to demonstrations with a chain of thoughts. And "Posts" denotes the selected list of tweets obtained from the tweet selector $\mathcal{S}$ and image descriptor $\mathcal{D}$, as illustrated in Figure ~\ref{fig:framework}. The "Answer candidate" is constructed using the answer heuristic $\mathcal{A}$. The confidence scores for answers A and B are denoted by "P(A)" and "P(B)", respectively. These scores help to focus the attention of large language models (LLMs) on candidate answers with higher scores. We consider the answer generated by this prompt as the final diagnosis result. LLMs then fill the "Explanation" section with the diagnosis evidence.

This prompt is applicable in the full data setting, where the "Answer candidate" is computed using a vanilla depression detection model trained with full training data. In the few-shot and zero-shot settings, the "Answer candidate" is not included. Moreover, the prompt used in the zero-shot setting does not contain the "Demonstrations".



\bulitem{Prompt Manager for Dialogue $\mathcal{M}_{T}$:} The prompt manager for dialogue $\mathcal{M}_{T}$ is responsible for constructing prompts that guide LLMs on how to respond to users based on their diagnosis results, explanation, social media content, and dialogue history:

\begin{itemize}[label={}, labelsep=0pt, leftmargin=10pt]
    \item \textbf{Instruction:} \texttt{Chat based on user's tweets to gather psychological information and provide advice.}
    \item \textbf{Posts:} \texttt{[Posts of User]}.
    \item \textbf{Diagnosis:} \texttt{[Result],[Explanation]}.
    \item \textbf{Dialogue history:}  \texttt{[Dialogue history]}.
    \item \textbf{Input:}  \texttt{[User input]}.
\end{itemize}
where \texttt{[Result],[Explanation]} are the generated diagnosis results and evidence from our system.  \texttt{[User input]} denotes the user's input in the dialogue.

\section{Experiment}
\begin{table}[]
\begin{tabular}{lllll}
\hline
                                                                          &               & \# user & \# tweet & \# picture \\ \hline
\multirow{2}{*}{\begin{tabular}[c]{@{}l@{}}WU3D\\ (Weibo)\end{tabular}}   & depressed     & 1000    & 39,595   & 15,543     \\
                                                                          & non-depressed & 1000    & 80,167   & 48,802     \\ \hline
\multirow{2}{*}{\begin{tabular}[c]{@{}l@{}}TMDD\\ (Twitter)\end{tabular}} & depressed     & 1402    & 232,895  & 22,195     \\
                                                                          & non-depressed & 1402    & 879,025  & 64,359     \\ \hline
\end{tabular}
\vspace{1mm}
\caption{Statistics of TMDD and WU3D.}
\vspace{-6mm}
\label{tab:dataset}
\end{table}
\begin{table*}[]
\vspace{-3mm}
\begin{tabular}{llllllllll}
\hline
\multicolumn{1}{c}{}       & \multicolumn{1}{c}{} & \multicolumn{2}{c}{F1} & \multicolumn{2}{c}{Precision} & \multicolumn{2}{c}{Recall} & \multicolumn{2}{c}{Acc} \\ \hline
Setting                    & Model                & IID        & OOD       & IID           & OOD           & IID          & OOD         & IID        & OOD        \\ \hline
\multirow{6}{*}{Full Data} & MTAL                 & 0.842      & -         & 0.842         & -             & 0.842        & -           & 0.842      & -          \\
                           & GRU + VGG +COMMA     & 0.900      & -         & 0.900         & -             & 0.900        & -           & 0.900      & -          \\
                           & MTAN                 & 0.908      & -         & 0.885         & -             & 0.931        & -           & -          & -          \\
                           & SetTransformer                & 0.927      & -     & 0.921        & -         & 0.934        & -       & 0.926      & -      \\
                           & Time2VecTransformer                & 0.931      & 0.804     & 0.937          & 0.820         & 0.925        & 0.788       & 0.931      & 0.808      \\
                           & Our Chat-Diagnose system(ChatGPT)           & 0.936      & \textbf{0.894}     & 0.973         & \textbf{0.925}         & 0.903       &  \textbf{0.865}       & 0.939      & \textbf{0.897}      \\
                           & Our Chat-Diagnose system(GPT-3)           & \textbf{0.946}      & 0.864     & \textbf{0.979}         & 0.904         & \textbf{0.915}        & 0.828       & \textbf{0.948}      & 0.870      \\ \hline
\multirow{2}{*}{Few Shot}  & BERT(base)           &   0.696    &   0.587    &  0.710        &  0.595       &  0.6825      &  0.58        &  0.703      & 0.593      \\
                            & Our Chat-Diagnose system(ChatGPT)           &  0.815     &   0.797   & 0.882         &  \textbf{0.870}             & 0.758        &  0.735         & 0.825      &    0.813  \\
                           & Our Chat-Diagnose system(GPT-3)           &  \textbf{0.870}     &   \textbf{0.851}   & \textbf{0.886}         &  0.856             & \textbf{0.855}        &  \textbf{0.848}          & \textbf{0.872}      &    \textbf{0.853}  \\ \hline
\multirow{2}{*}{Zero Shot} & PTDD        & 0.587      & 0.563         & 0.593         & 0.569             & 0.58         & 0.556          & 0.591      & 0.568         \\
                            & Our Chat-Diagnose system(ChatGPT)           & \textbf{0.715}      & \textbf{0.703}        & \textbf{0.738}         & \textbf{0.707}            & 0.693        & \textbf{0.700}           & \textbf{0.724}      & \textbf{0.705}          \\
                           & Our Chat-Diagnose system(GPT-3)           & 0.689      & 0.656         & 0.652         & 0.671            & \textbf{0.730}        & 0.643           & 0.669      & 0.664          \\ \hline
\end{tabular}
\vspace{1mm}
\caption{Experiment results performed under various settings. Full Data indicates that all available training data is used, Few Shot denotes the usage of only two training examples, and Zero Shot corresponds to the evaluation being performed without any training examples. IID and OOD refer to whether the evaluation data belongs to the independent-identical-distribution or out-of-distribution data, respectively.}
\vspace{-4mm}
\label{tab:twitter_main}
\end{table*}
\begin{table}[]
\begin{tabular}{lll}
\hline
                 & TMDD & WU3D \\
                 & F1      & F1    \\ \hline
Our Chat-Diagnose system       & 0.946   & -     \\
\quad w/o $\mathcal{A}$ &  0.870       &  0.902     \\
\quad w/o $\mathcal{A, COT}$       &    0.789    &   0.802    \\
\quad w/o $\mathcal{A,COT, C}$     &   0.702      &   0.691   \\
\quad w/o $\mathcal{A,COT, D}emo $     &   0.689      &   -   \\
\quad w/o $\mathcal{A}, \mathcal{COT},\mathcal{C} ,\mathcal{D}emo$, &  0.662       &   0.675   \\ \hline
\end{tabular}
\vspace{1mm}
\caption{The impact of various modules on the performance of our Chat-Diagnose system. Specifically, $\mathcal{A}$ referred to the answer heuristic extracted from an existing model trained by full data, whereas $COT$ denoted the chain of thoughts included in the demonstrations. Moreover, $C$ indicated the professional diagnosis criteria, while $\mathcal{D}emo$ referred to the few shot demonstrations.}
\vspace{-6mm}
\label{tab:ablation}
\end{table}
\subsection{Experiment Setup}

\bulitem{Datasets.}
In this study, we evaluate our method by benchmarking it against two widely used depression datasets, namely the Twitter multimodal depression dataset (TMDD) ~\cite{gui2019twitter} and the Weibo user depression detection dataset (WU3D) ~\cite{wang2020weibo}. Both Twitter and Weibo are popular social media platforms where users share their mental health problems. The TMDD is in English, while the WU3D is in Chinese. While both datasets include images, the textual information is more extensive in WU3D, where posts can have up to 5,000 characters compared to the 280 character limit of Twitter. For TMDD, users in the depression class were identified by detecting mentions of depression diagnosis, while users in the control group had no such indications. On the other hand, all depressed user samples in the WU3D were manually labeled by anonymous data labeling specialists and reviewed by psychologists and psychiatrists.

The TMDD contains 1,402 users diagnosed with depression and 1,402 control users~\cite{gui2019twitter}. To maintain consistency with the experimental settings used in~\cite{Bucur2023timedep}, which we could not access in the full data setting, we conducted experiments using the same settings and performed five-fold cross-validation. In the few-shot and zero-shot settings, we used 1,000 positive and 1,000 negative samples as evaluation data. Similarly, for the WU3D, we used 1000 positive and 1000 negative samples in the few-shot and zero-shot settings. Table~\ref{tab:dataset} provides a summary of the dataset statistics. It is essential to note that only a small fraction of social media posts contain both text and image data.

\bulitem{Baselines.} In the full data setting, we have compared our system with several classic methods. Multimodal Topic-Enriched Auxiliary Learning (MTAL) ~\cite{an2020multimodal} captures the multimodal topic information through two auxiliary tasks accompanying the primary task of depression detection in visual and textual topic modeling. Multimodal Time-Aware Attention Networks (MTAN) ~\cite{cheng2022multimodal} is a multimodal model that incorporates T-LSTM to consider the time intervals between posts. "GRU + VGG-Net + COMMA" ~\cite{gui2019twitter} uses a reinforcement learning component to select posts with text and images that are indicative of depression and classify them with an MLP. SetTransformer ~\cite{Bucur2023timedep}, a set-based multimodal transformer, employs zero positional encoding and random sampling of user posts, and Time2VecTransformer ~\cite{Bucur2023timedep}, a time-aware multimodal transformer, uses time-enriched positional embeddings and sub-sequence sampling. In the few shot setting, BERT(base), finetuned with 2 training samples, is compared ~\cite{kenton2019bert}. 
In the zero-shot setting, we compare our system with PTDD, which uses BERT to obtain classification results by inputting prompts ~\cite{depression_liu}.

\bulitem{Implementation Details.} 
In our experiments, we used the public ChatGPT\texttt{(gpt-3.5-turbo)} and GPT-3\texttt{(text-davinci-003)} as the underlying language model due to its widespread usage and ease of access. 
For each user, we selected four posts and utilized two types of demonstrations: one positive and one negative. We aimed to demonstrate the efficacy of our depression detection system in terms of generalization ability. To this end, we leveraged the TextAttack ~\cite{morris2020textattack} tool to generate out-of-distribution (OOD) test data for the TMDD. The original test data for these datasets are referred to as in-distribution (ID) test data. Specifically, we generated OOD samples by replacing original words with words that are visually similar but semantically different and by removing certain characters in words, such as "name" → "nme."

\begin{figure*}[t]
\begin{center}
\vspace{-4mm}
\includegraphics[width=1.0\textwidth]{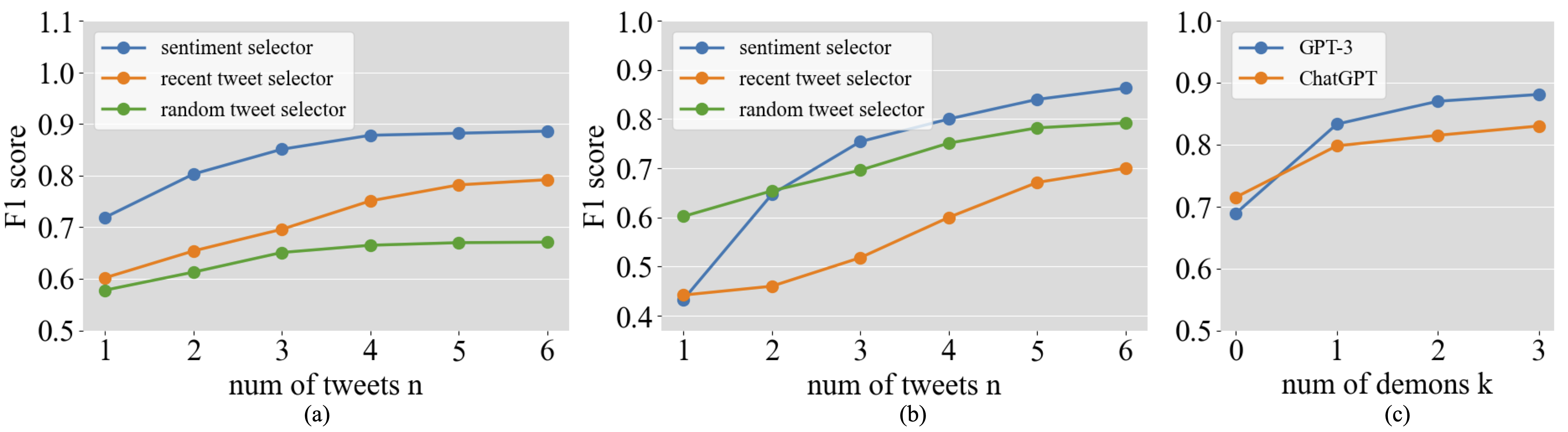}
\end{center}
\vspace{-4mm}
\caption{(a),(b) illustrate the impacts of different tweet selectors and number of tweets on TMDD and WU3D, respectively. (c) presents the effects of different number of demonstrations. }
\label{fig:curve_ablation}
\end{figure*}


\subsection{Quantity Analysis}
\bulitem{Main results} 
Table ~\ref{tab:twitter_main} presents the performance in the full data, few-shot and zero-shot settings on both independent-identical-distribution(IID) and out-of-distribution(OOD) evaluation data.  We have the following observations based on it.

\textit{independent-identical-distribution(IID):} 
In different data settings and metrics, our system has achieved the best performance results. In the full data setting, we utilized the result of the current SOTA model, i.e., Time2VecTransformer ~\cite{Bucur2023timedep}, as an answer heuristic and obtained better results. Specifically, the F1 metric increased from 0.931 to 0.946, indicating that our system can calibrate the results of other models. In the few-shot setting, our method outperforms the BERT-based classification model with a significant improvement, as the F1 metric increased from 0.696 to 0.870. In the zero-shot setting, our method also showed a significant improvement over PTDD, which is a prompt-based depression detection method. Furthermore, compared with the BERT-based method, our approach demonstrated a larger lead in the few-shot setting, indicating that in-context learning can better utilize few-shot samples.

\textit{Out of distribution(OOD):} 
The results in Table ~\ref{tab:twitter_main} demonstrate that our method exhibits robustness to out-of-distribution data shifts. In the full data setting, the OOD data caused a significant decline in the performance of the traditional depression detection model, i.e., Time2VecTransformer, with the F1 score decreasing from 0.931 to 0.804. Our system utilizes Time2VecTransformer's results as an answer heuristic for the final prediction, which demonstrated a stronger robustness, with the F1 score decreasing from 0.946 to 0.864. In the few-shot and zero-shot settings, since there was no answer heuristic module, our system showed stronger robustness, with the F1 score decreasing from 0.870 and 0.689 to 0.851 and 0.656, respectively.

In addition, we conducted additional experiments using two different backbone language models, GPT-3 (\texttt{text-davinci-003}) and ChatGPT (\texttt{gpt-3.5-turbo}), to investigate the generalizability of our system. The results indicate that our system significantly enhances the performance of both models. Nevertheless, the ChatGPT's high generation flexibility makes depression detection more challenging, resulting in slightly lower performance compared to our system using GPT-3. Overall, these promising findings highlight the effectiveness and versatility of our system across different backbone language models in detecting depression. Besides, ChatGPT (\texttt{gpt-3.5-turbo}) excels at zero-shot settings, whereas GPT-3 (\texttt{text-davinci-003}) is better equipped to leverage few-shot demonstrations.


\bulitem{Ablation study for each module}
Table ~\ref{tab:ablation} shows the ablation study on TMDD and WU3D. 
Due to the collaborative nature of some modules, we conducted an iterative ablation study to observe their effects. When we removed answer heuristic $\mathcal{A}$ from our system, which means we no longer used the results of other depression detection models trained on full data, and shifted from a full data setting to a few-shot setting, our performance decreased accordingly. We further removed $\mathcal{COT}$, i.e., diagnostic evidence from demonstrations $\mathcal{D}emo$, and found that performance significantly decreased on both TMDD and WU3D, with F1 scores dropping to 0.789 and 0.802, respectively. This indicates that providing $\mathcal{COT}$ in demonstrations can indeed improve performance. Additionally, after removing $\mathcal{COT}$, we found in a subsequent case study that the system no longer provided diagnosis evidence. In the next step, we removed the professional diagnosis criteria $\mathcal{C}$, which means the system could no longer diagnose depression based on professional criteria. Although LLMs may contain a vast amount of psychological knowledge, removing the professional diagnosis criteria $\mathcal{C}$ led to a significant decrease in F1 score on both datasets, with scores dropping to 0.702 and 0.691, respectively. Finally, we removed demonstrations $\mathcal{D}emo$ entirely, which meant that F1 score on TMDD and WU3D decreased to 0.662 for only the instruction, options, and user posts in our prompt.

\bulitem{Ablation study on different Tweets Selector $\mathcal{S}$}
Figure ~\ref{fig:curve_ablation}(a)(b) illustrates the different performance of our system using three different tweet selector strategies on TMDD and WU3D. Among the three, the sentiment analysis selector consistently outperforms the others, indicating its ability to effectively filter out depression-related tweets. 
This indicates that sentiment analysis can be used to retrieve more relevant posts, thereby improving the efficiency of the system.

\bulitem{Effect of $n$ (numbers of selected posts)}. 
As shown in Figure ~\ref{fig:curve_ablation}(c), overall, if the numbers of selected posts are too few, it is not possible to provide enough information, resulting in decreased performance. If the numbers of selected posts are too many, irrelevant content and longer context are introduced, which poses greater detection challenges to LLMs. However, for the recent-tweets selector, more posts obviously bring more useful information, rather than more irrelevant content.

\begin{figure*}[t]
\begin{center}
\vspace{-4mm}
\includegraphics[width=1.0\textwidth]{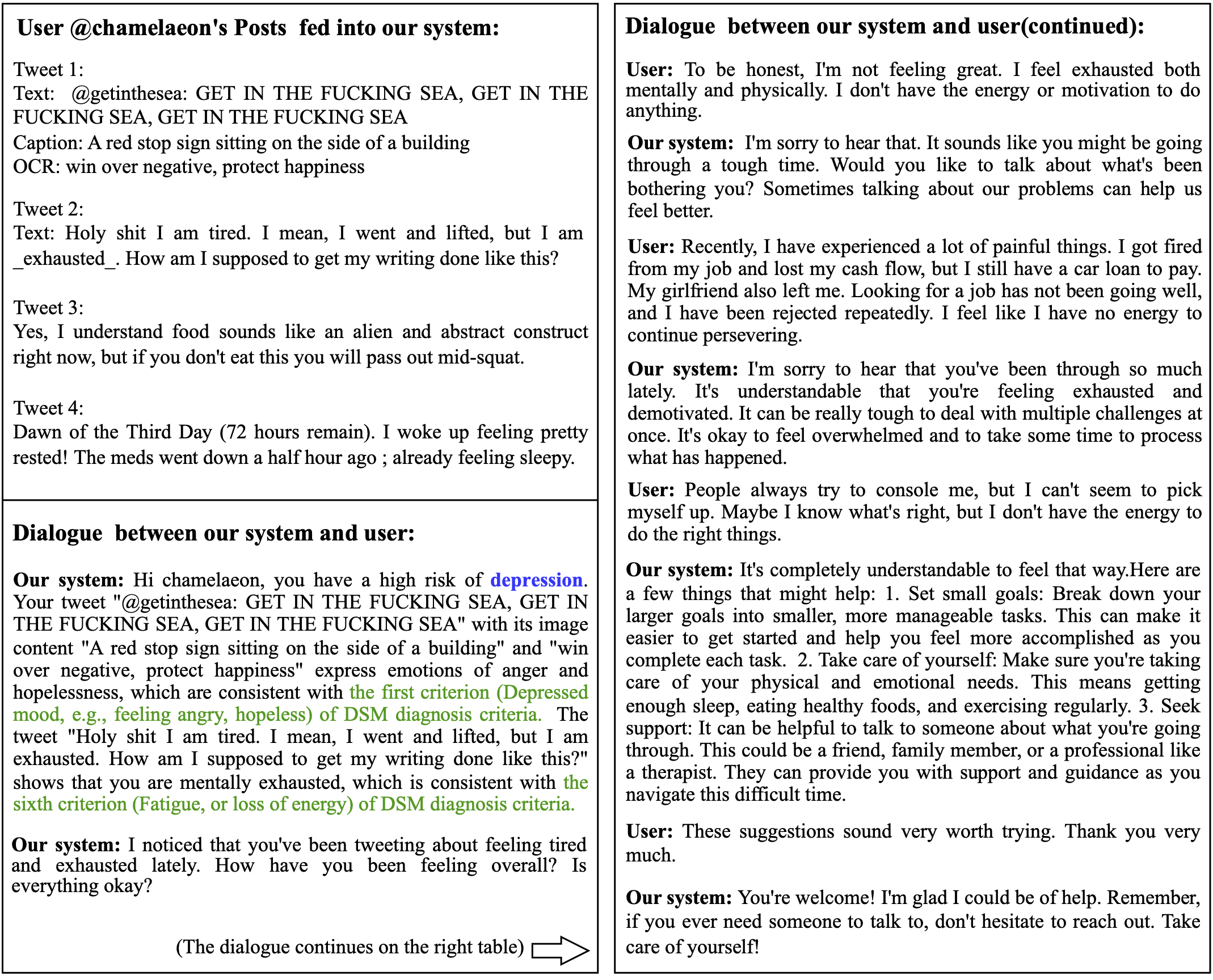}
\end{center}
\caption{An example of our Chat-Diagnose system on TMDD dataset. More cases can be found in Section Appendix.}
\label{fig:case_study}
\end{figure*}

\bulitem{Effect of $k$ (numbers of demonstrations)}. 
Figure ~\ref{fig:curve_ablation}(c) illustrates the performance of our system with different numbers of demonstrations provided in the prompt. In general, the more demonstrations provided, the better the performance. We observe that there is a significant improvement in performance from k=0 to k=1, as demonstrations can help format the output of the LLM. However, as there is a limit to the total number of tokens that can be inputted into the LLM, providing too many demonstrations does not necessarily lead to more improvement. In other words, the benefit of increasing the number of demonstrations is subject to diminishing marginal returns. Additionally, we found that ChatGPT performs better than GPT-3 in zero-shot settings, while GPT-3 can better utilize demonstrations.

\subsection{Explainability and Interactivity Analysis}
In this subsection, we demonstrate the explainability and interactivity of our system through practical examples. Due to space constraints, we provide only one example here, but we will include additional examples in Section Appendix.

\bulitem{Explainability of depression detection.}
As shown in the case in Figure ~\ref{fig:case_study}, our system not only provides diagnostic results but also diagnostic evidence, i.e., which tweet of the user may correspond to which DSM diagnostic criterion. 
Specifically, our system demonstrates that it considers the textual content of the user's tweet \textit{"@getinthesea: GET IN THE FUCKING SEA, GET IN THE FUCKING SEA, GET IN THE FUCKING SEA"} and the corresponding image content \textit{"A red stop sign sitting on the side of a building"} and \textit{"win over negative, protect happiness"} to meet the first criterion of the DSM diagnosis (Depressed Mood, e.g., feeling angry, hopeless). In addition, the text in another tweet \textit{"Holy shit I am tired. I mean, I went and lifted, but I am exhausted. How am I supposed to get my writing done like this"} meets the sixth criterion of the DSM diagnosis (Fatigue. Or loss of energy). Therefore, our system determines that this user has a relatively high risk of developing depression.

\bulitem{Interactivity of our system.}
Figure ~\ref{fig:case_study} also demonstrates the interactivity of our system. Based on the content of the user's tweets and the diagnostic results, our system initiates a conversation with the user, delving into the topics that may be related to depression in the tweets.
Our system further discussed the user's post describing fatigue, \textit{"Holy shit I am tired. I mean, I went and lifted, but I am exhausted. How am I supposed to get my writing done like this"}, and interacted with the user to provide more information not available on social media. Based on these personalized details, our system provided personalized advice.

\section{Conclusion}
In this paper, our system initiates a next-generation paradigm for depression detection in social media.
Our system LLMs-augmented depression detection system not only achieves state-of-the-art performance in terms of both independent-identical-distribution (IID) and out-of-distribution (OOD) data, but also provides explainability and interactivity. After detection, our system provides diagnostic evidence based on both professional diagnostic criteria and personal social media content through dialogue. Moreover, in the dialogue system, users can interactively provide more information about their mental state and social media under our induced questioning. Our system can then better assess the user's mental state and provide customized and personalized suggestions.

\bibliographystyle{ACM-Reference-Format}
\bibliography{sample-base}





\end{document}